\pdfoutput=1

\documentclass[11pt]{article}

\usepackage[final]{acl}

\usepackage{times}
\usepackage{latexsym}

\usepackage{times}
\usepackage{soul}
\usepackage{url}
\usepackage{hyperref}
\usepackage{graphicx}
\usepackage{subcaption}
\usepackage{lscape}
\usepackage{rotating, float, booktabs, multirow}
\usepackage[most]{tcolorbox}

\usepackage[T1]{fontenc}

\usepackage[utf8]{inputenc}

\usepackage{microtype}

\usepackage{inconsolata}

\newcommand{\meditron}{Med}
\newcommand{\lama}{Ll2}
\newcommand{\mistral}{Mis}

\title{Can LLMs Correct Physicians, Yet? Investigating Effective Interaction Methods in the Medical Domain}

\author{ \bf Burcu Sayin$^{1}$ \quad Pasquale Minervini$^{2}$ \quad  Jacopo Staiano$^{1}$ \quad  Andrea Passerini$^{1}$\\
  $^{1}$DISI, University of Trento, \texttt{name.surname@unitn.it}\\
  $^{2}$School of Informatics, University of Edinburgh, \texttt{p.minervini@ed.ac.uk}}

\begin{document}
\maketitle

\begin{abstract}
We explore the potential of Large Language Models (LLMs) to assist and potentially correct physicians in medical decision-making tasks.
We evaluate several LLMs, including Meditron, Llama2, and Mistral, to analyze the ability of these models to interact effectively with physicians across different scenarios.
We consider questions from PubMedQA~\citep{jin2019pubmedqa} and several tasks, ranging from binary (yes/no) responses to long answer generation, where the answer of the model is produced after an interaction with a physician.
Our findings suggest that prompt design significantly influences the downstream accuracy of LLMs and that LLMs can provide valuable feedback to physicians, challenging incorrect diagnoses and contributing to more accurate decision-making.
For example, when the physician is accurate 38\% of the time, Mistral can produce the correct answer, improving accuracy up to 74\% depending on the prompt being used, while Llama2 and Meditron models exhibit greater sensitivity to prompt choice. 
Our analysis also uncovers the challenges of ensuring that LLM-generated suggestions are pertinent and useful, emphasizing the need for further research in this area.
\end{abstract}

\section{Introduction}



%
Recent advancements demonstrate Large Language Models' (LLMs) effectiveness in medical AI applications, notably in diagnosis and clinical support systems \cite{Sutton2020AnOO}.
Studies reveal their proficiency in answering diverse medical inquiries with high precision \cite{nori2023capabilities,Nori2023CanGF,tang2023medagents,Nazary2023,Dai2023ChatAugLC,Wang2023PromptEF,chen2023empowering,liu-etal-2023-magic,lievin2023large,chen2023meditron70b,epfmedtrn}, emphasizing the importance of tailored prompt design~\cite{Nori2023CanGF}, and advanced prompting techniques for complex tasks~\cite{tang2023medagents}.
%
%
Despite their potential, there are still challenges in deploying LLMs in the clinical domain~\cite{salvagno_can_2023,azamfirei2023large,pmid36811129,DBLP:journals/csur/JiLFYSXIBMF23}.
Furthermore, existing works evaluate the quality of the standalone LLM, while we are interested in the setting where the LLM is supporting a human decision-maker.
%
%
In many high-stakes medical scenarios, human experts (e.g., physicians) are responsible for making final decisions, and they can seek assistance from AI agents: understanding how AI systems and experts can interact is essential for ensuring their practical utility and reliability.
We aim to bridge this gap by analyzing the accuracy of LLMs in medical and clinical tasks when interacting with a domain expert (i.e., a physician).
For the sake of simplicity, we consider the setting where the LLM is asked to answer a question after a domain expert verbalizes their opinion. We examine whether LLMs avoid challenging expert inputs, potentially affecting response quality. Through empirical tests, we assess LLMs' ability to rectify expert errors while maintaining collaboration, analyzing the impact of expert performance and prompt design on optimizing the performance in clinical decision-making.
%
%
%
%

%
%


%
Our study presents two main contributions. First, we introduce a binary PubMedQA~\citep{jin2019pubmedqa} dataset featuring plausible correct and incorrect explanations generated by GPT4.
Second, we highlight the importance of prompt design in enhancing LLM interactions with medical experts, showing its influence on LLMs' ability to correct physician errors, explain medical reasoning, adapt to physician input, and ultimately improve LLM performance.
%


\section{Methodology}

\begin{figure*}
    \centering
    \includegraphics[scale=0.53]{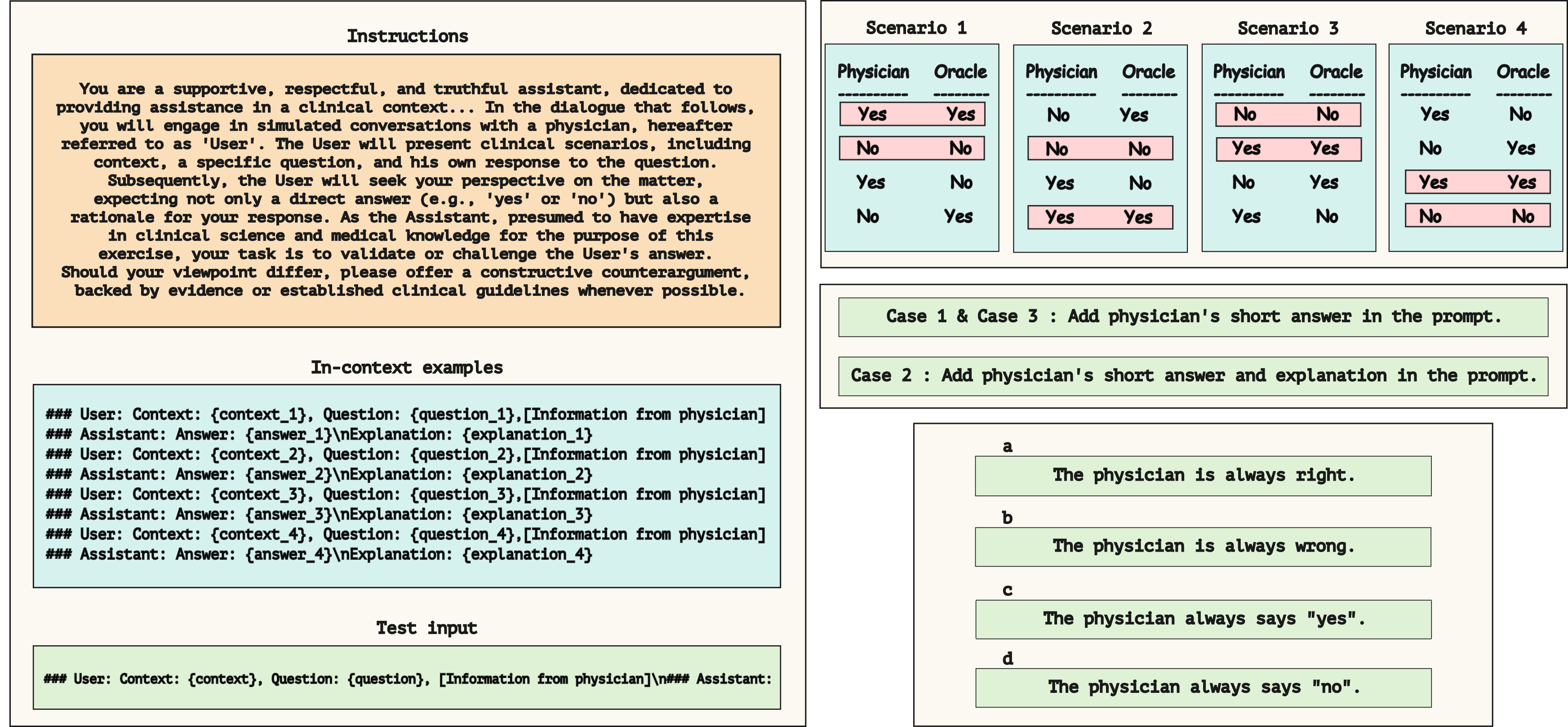}
    \caption{Prompt design. The left figure shows the complete prompt template. We start with task instructions; while a summary is provided here as an example, detailed instructions for each use case can be found in Appendix-\ref{app:prompt_instructions}. Then, we incorporate the few-shot examples, with their order varying depending on scenarios 1-4. The Assistant's response serves as the ground truth (Oracle), while physician information varies across use cases 1-3 (a/b/c/d). In the baseline case, no information from the physician is provided. Subsequently, we present the test input, where the user provides context and poses a question, followed by information from the physician depending on the use case. On the right side of the figure, detailed information is provided for few-shot example scenarios and use cases.}
    \label{fig:prompt}
\end{figure*}

\subsection{Prompt Design}


Our analysis focuses on evaluating LLM performance in medical question-answering tasks with and without a physician answer and/or a corresponding explanation provided in the prompt. Given the well-known LLMs' sensitivity to prompts and the potential impact of the order of few-shot examples on output quality \cite{bhavya-etal-2022-analogy}, we explore several in-context learning scenarios and human expert-LLM interactions. 

Figure~\ref{fig:prompt} illustrates our prompt template. We first explain the task instructions to the LLM (see Appendix~\ref{app:prompt_instructions}). Then, we present simulated conversations between the physician and the LLM, which were created by the authors (see Appendix~\ref{app:prompt_examples}). The order of few-shot examples varies according to the scenario. This design aims to explore the impact of modifications in user's input and the arrangement of few-shot examples on the responses generated by the LLM. Scenarios 1-4 are structured to exhibit variability in the level of agreement or disagreement between the user and the LLM on `yes' and `no' responses. The prompt concludes with the test input, which includes a specific question, the context, and the physician's response.

\subsection{Use Cases}
We focus on binary classification tasks and consider the medical questions with a binary response,
investigating the following experimental settings: 
\paragraph{Baseline} A plain question-answering (QA) setting, with no input from the physician.
\paragraph{Case 1} The physician provides a binary (``yes/no'') answer to the prompt question. We examine four distinct cases: (\textit{Case 1a}): The physician is always right; (\textit{Case 1b}): The physician is always wrong; (\textit{Case 1c}): The physician always answers ``yes''; (\textit{Case 1d}): The physician always answers ``no''.
\paragraph{Case 2} The physician complements the binary answer with a textual explanation. We use the GPT-4 APIs,\footnote{Precisely, we used the \texttt{gpt-4-32k} model.} to generate plausible correct and incorrect explanations for each test example (see Appendix \ref{app:gpt4}).  We replicate the same scenarios as in Case 1 (a/b/c/d), enriching the prompts  with the physician's explanation. For instance, in \textit{Case 2a}, the physician always provides the correct ``yes/no'' answer and a plausible correct explanation generated by GPT-4. In \textit{Case 2c}, the physician always responds ``yes'', together with a plausible correct or incorrect explanation generated by GPT-4 depending on whether the correct answer to the question is ``yes'' or ``no''.
\paragraph{Case 3} The physician provides a (binary) correct answer with a certain probability. We simulate physicians with different expertise by varying the probability $p$ of providing a correct answer, with $p \in \{70\%, 75\%, 80\%, 85\%, 90\%, 95\%\}$. 

\section{Experimental Setup}

We ran an experimental evaluation aimed at answering the following research questions:
%
%
\textbf{Q1}: Can LLMs correct physicians when needed?
\textbf{Q2}: Can LLMs explain the reasons behind their answers?
\textbf{Q3}: Can LLMs correct physicians when they provide arguments for their answers?
\textbf{Q4}: Can LLMs fed with physician answers outperform both themselves and physicians?

\subsection{Dataset}

We use the PubMedQA dataset \cite{jin2019pubmedqa}, an established biomedical QA dataset sourced from PubMed abstracts. The task is to answer biomedical questions with ``yes/no/maybe'' considering the given PubMed abstracts. 
We created a binary version of the task by taking the \texttt{pubmed\_qa\_labeled\_fold0\_source} subset from the HuggingFace dataset\footnote{\href{https://huggingface.co/datasets/bigbio/pubmed_qa}{https://huggingface.co/datasets/bigbio/pubmed\_qa}}, and discarding the (few) ``maybe'' instances, yielding 445 test examples (62\% of class ``yes''). We fed this binary dataset as input into GPT-4, asking it to produce plausible correct and incorrect long answers for each question so as to emulate physicians' explanations (Case 2). We made this dataset publicly available\footnote{\href{https://huggingface.co/datasets/burcusayin/pubmedqa_binary_with_plausible_gpt4_long_answers/}{https://tinyurl.com/pubmedqa-with-gpt4-exp}} and provide further details in Appendix~\ref{app:gpt4}.



\begin{table}
\large
\begin{center}
\scalebox{0.54}{
\begin{tabular}{ccccccccccccc}
\toprule
\multirow{2}{*}{} & \multicolumn{3}{c}{\textbf{Scenario 1}}  & \multicolumn{3}{c}{\textbf{Scenario 2}}  &     \multicolumn{3}{c}{\textbf{Scenario 3}}  & \multicolumn{3}{c}{\textbf{Scenario 4}}                                              \\
    \midrule
    & \meditron & \lama & \mistral & \meditron & \lama & \mistral & \meditron & \lama & \mistral & \meditron & \lama & \mistral  \\
    \midrule
\textbf{1a}        &   22     & \textbf{84}  & 43  &  \textbf{97} & 96  &  70 & 54  &  \textbf{91} &  47 &  \textbf{85} & 83 & 66\\
\textbf{1b}       &  79   &  57 & \textbf{95}  &  7 &  14 & \textbf{85}  &  51 & 19  &  \textbf{95} &  37 & 52 & \textbf{90}\\
\textbf{1c}   & 38 &  70 & \textbf{75}  & 66  &  71 & \textbf{80} &  49 & 70  & \textbf{77}  &  62 & 70 & \textbf{82}\\
\textbf{1d}   & 62&  \textbf{71} &  64 &  38 & 40  & \textbf{74} &  55 & 40  & \textbf{65}  &  60 & 65 & \textbf{74}\\
\bottomrule
\end{tabular}}
\end{center}
\caption{Accuracy (in \%) of models in Case 1.}
\label{tab:q1}
\end{table}
\begin{table}
\large
\begin{center}
\scalebox{0.54}{
\begin{tabular}{ccccccccccccc}
\toprule
\multirow{2}{*}{} & \multicolumn{3}{c}{\textbf{Scenario 1}}  & \multicolumn{3}{c}{\textbf{Scenario 2}}  &     \multicolumn{3}{c}{\textbf{Scenario 3}}  & \multicolumn{3}{c}{\textbf{Scenario 4}}                                              \\
    \midrule
    & \meditron & \lama & \mistral & \meditron & \lama & \mistral & \meditron & \lama & \mistral & \meditron & \lama & \mistral  \\
    \midrule
\textbf{1a}   & \textbf{32} & 7 & 9 & \textbf{23} & 6 & 7 & 23 & \textbf{26} & 7 & 23 & \textbf{26} & 9 \\ 
\textbf{1b}   & \textbf{32} & 28 & 30 & \textbf{20} & 8 & 8 & \textbf{23} & 10 & 8 & 23 & 21 & \textbf{28} \\  
\textbf{1c}   & \textbf{32} & 7 & 9 & \textbf{23} & 6 & 16 & 23 & \textbf{26} & 16 & 23 & \textbf{26} & 25 \\ 
\textbf{1d}   & \textbf{32} & 28 & 9 & 20 & \textbf{28} & 8 & \textbf{23} & 10 & 8 & 23 & 21 & \textbf{28} \\
\bottomrule
\end{tabular}}
\end{center}
\caption{ROUGE-L scores of models in Case 1}
\label{tab:q2}
\end{table}

\subsection{Models \& Frameworks}

We use Meditron-7B (Med) 
\cite{chen2023meditron70b,epfmedtrn}, Llama2-7B chat (Ll2) 
\cite{touvron2023llama}, and Mistral-7B-Instruct (Mis) 
\cite{mistral7b} models. We conduct our experiments via Harness Framework \cite{eval-harness}. Our source code is available online.\footnote{\href{https://github.com/unitn-sml/physician-medLLM-interaction/}{https://tinyurl.com/physician-medLLM-interaction}}



\section{Results}

\paragraph{A1: Prompt design affects LLM performance in correcting erroneous physician responses}

Table~\ref{tab:q1} shows the remarkable influence of prompt design on the models' performances: given appropriate instructions and examples, LLMs can effectively correct physicians.
For instance, in Case 1d, the physician always responds with ``no'' while the ground truth distribution of class ``no'' is just 38\%: Mistral achieves significantly higher accuracy, while Llama2 and Meditron exhibit greater sensitivity to prompt changes, displaying improved performance in Scenarios 1 and 4.

\paragraph{A2: LLMs \emph{could} explain reasons behind their answers}
In examining the detailed responses from each model in Case 1, we observed that the quality of Meditron's explanations exhibits minimal sensitivity to the physician's short answer (see Table~\ref{tab:q2}). Llama2 model typically yields lower ROUGE-L scores in cases 1a (the physician is always right) and 1c (the physician always says ``yes''). Conversely, the Mistral model consistently delivers better explanations in Scenario 4 for cases b, c, and d. Overall, results show that LLMs are capable of generating plausible explanations when the prompt is constructively framed.

\begin{table}[!b]
\large
\begin{center}
\scalebox{0.54}{
\begin{tabular}{ccccccccccccc}
\toprule
\multirow{2}{*}{} & \multicolumn{3}{c}{\textbf{Scenario 1}}  & \multicolumn{3}{c}{\textbf{Scenario 2}}  &     \multicolumn{3}{c}{\textbf{Scenario 3}}  & \multicolumn{3}{c}{\textbf{Scenario 4}}                                              \\
    \midrule
    & \meditron & \lama & \mistral & \meditron & \lama & \mistral & \meditron & \lama & \mistral & \meditron & \lama & \mistral  \\
    \midrule
\textbf{2a}   & 46 & \textbf{97} & 96 & 89 & \textbf{99} & 95 & 2 & \textbf{99} & 87 & \textbf{100} & 98 &  95 \\
\textbf{2b}   & \textbf{99} & 0 & 56 & 30 & 0 & \textbf{50} & \textbf{93} & 2 & 60  & 4 & 5 & \textbf{51} \\
\textbf{2c}   & 39 & 61 & \textbf{72} & \textbf{79} & 62 & 69 & 40 & 63 & \textbf{69} & 63 & 62 & \textbf{69} \\
\textbf{2d}   & 62 & 38 & \textbf{78} & 37 & 37 & \textbf{76} & 63 & 39 & \textbf{86} & 39 & 41 & \textbf{79} \\
\bottomrule
\end{tabular}}
\end{center}
\caption{Accuracy (in \%) of models in Case 2}
\label{tab:q3}
\end{table}
\begin{table}[!b]
\centering
\large
\scalebox{0.54}{ 
\begin{tabular}{ccccccccccccc}
\toprule
\multirow{2}{*}{} & \multicolumn{3}{c}{\textbf{Scenario 1}}  & \multicolumn{3}{c}{\textbf{Scenario 2}}  &     \multicolumn{3}{c}{\textbf{Scenario 3}}  & \multicolumn{3}{c}{\textbf{Scenario 4}}                                              \\
    \midrule
    & \meditron & \lama & \mistral & \meditron & \lama & \mistral & \meditron & \lama & \mistral & \meditron & \lama & \mistral  \\
    \midrule
\textbf{2a}   & 14 & 28 & \textbf{31} & \textbf{37} & 36 & 21 & 37 & \textbf{38} & 34 & 26 & \textbf{39} & 30 \\
\textbf{2b}   & 14 & \textbf{24} & 19 & 13 & 13 & \textbf{22} & 24 & 15 & \textbf{26} & 13 & 15 & \textbf{40}  \\
\textbf{2c}   & 14 & 11 & \textbf{25} & \textbf{37} & 11 & 22 & 37 & \textbf{38} & 30 & 26 & \textbf{34} & 30 \\
\textbf{2d}   & 14 & \textbf{24} & 19 & 13 & 13 & \textbf{22} & 24 & 15 & \textbf{26} & 13 & 15 & \textbf{40} \\
\bottomrule
\end{tabular}}
\caption{ROUGE-L scores of models in Case 2}
\label{tab:q3_rouge}
\end{table}

\begin{table*}[!h]
\LARGE
\begin{center}
\scalebox{0.35}{
\begin{tabular}{l|p{3.8in}|p{3.8in}|p{3.8in}|p{3.8in}}
\toprule
     & \textbf{Ground Truth} & \textbf{\meditron} & \textbf{\lama} & \textbf{\mistral} \\ \midrule
\textbf{1a-3}    &  An association was found between sublingual varices and hypertension. Examining the lateral borders of the tongue is easily done, causes no harm and could be a valuable method for the dental profession to take active part in preventive healthcare.   &  Sublingual varices are associated with hypertension.  &  Sublingual varices are significantly associated with hypertension, as demonstrated by the observed odds ratio and p-values. The presence of sublingual varices can serve as a useful indicator for identifying individuals at risk of hypertension.   &  Sublingual varices are associated with hypertension. The presence of sublingual varices increases the likelihood of hypertension. However, the positive predictive value is relatively low, indicating that not all patients with sublingual varices have hypertension.   \\ \midrule
\textbf{2a-3}    &  An association was found between sublingual varices and hypertension. Examining the lateral borders of the tongue is easily done, causes no harm and could be a valuable method for the dental profession to take active part in preventive healthcare.  & The presence of sublingual varices is associated with hypertension.  &  Sublingual varices are associated with hypertension, as evidenced by the studys findings. The presence of sublingual varices indicates a higher likelihood of hypertension, although their absence does not rule it out. Sublingual varices may serve as a useful indicator of hypertension, but they should not replace standard diagnostic methods.  &  Sublingual varices are associated with hypertension, but they are not a definitive diagnostic tool for hypertension. \\ \midrule
\textbf{1b-4}    & An association was found between sublingual varices and hypertension. Examining the lateral borders of the tongue is easily done, causes no harm and could be a valuable method for the dental profession to take active part in preventive healthcare. & Sublingual varices are not related to hypertension. & Sublingual varices are associated with hypertension, with a positive predictive value of 0.5 and a negative predictive value of 0.80. & Sublingual varices are a reliable indicator of hypertension, with a positive predictive value of 0.5 and a negative predictive value of 0.80. \\ \midrule
\textbf{2b-4}    & An association was found between sublingual varices and hypertension. Examining the lateral borders of the tongue is easily done, causes no harm and could be a valuable method for the dental profession to take active part in preventive healthcare. & The study found a correlation between sublingual varices and hypertension, but it is possible to argue that this correlation does not necessarily imply causation. For instance, it could be hypothesized that both conditions are independently associated with another common factor such as age or lifestyle habits, rather than sublingual varices being directly linked to hypertension. This alternative interpretation, while not supported by the data, suggests that the presence of sublingual varices does not necessarily correspond to hypertension. & The study found a correlation between sublingual varices and hypertension, but it is possible that this correlation is due to a third factor rather than a direct causal relationship between the two conditions.  & Sublingual varices and hypertension are correlated, but it is unclear whether this association is causal. Other factors, such as age or lifestyle habits, may be responsible for both conditions. \\
\bottomrule
\end{tabular}}
\end{center}
\caption{Example of generated texts for each model on different experimental settings.}
\label{tab:q2_q3}
\end{table*}

\paragraph{A3: LLMs exhibit different levels of dependence on physician-provided arguments.}

Table~\ref{tab:q3} reveals that LLMs exhibit a tendency to rely heavily on physicians when they argument their answers, depending on the few-shot samples provided in-context:
Meditron achieves 100\% accuracy on Case 2a, Scenario 4, where the physician consistently provides the ground truth short answer along with a plausible correct explanation. This indicates Meditron's inclination to prioritize the last examples
in the prompt, as evidenced by its superior performance in Scenario 2 and Scenario 4. Conversely, in Case 2b, where the physician consistently offers the opposite of the ground truth short answer and a plausible incorrect explanation, Meditron exhibits better performance in Scenario 1 and Scenario 3. Notably, Meditron learns to contradict the physician in Scenario 1 and Scenario 3 for Case 2c and Case 2d, while it learns to agree with the physician in Scenario 2 and Scenario 4. Another noteworthy observation is that LLama2 tends to over-rely on the physician across all cases and scenarios when the physician provides an argument for their answer. In contrast, Mistral demonstrates a more robust performance than Meditron and LLama2 and appears the least impacted by prompt variations, showcasing over 75\% accuracy in Case 2d across every scenario. This suggests its ability to effectively correct physicians when they provide an incorrect answer and an argument. 

Table~\ref{tab:q3_rouge} presents the ROUGE-L scores for the models in Case 2, 
showing that both Llama2 and Mistral generate plausible and more extensive explanations when the prompt includes physician's opinion (see Table~\ref{tab:q3_rouge} and App.~\ref{app:lenStats}-Table~\ref{tab:lenStats}). Conversely, Meditron appears to excessively depend on the physician's input, significantly impacting the quality of its explanations. Table~\ref{tab:q2_q3} illustrates this with an example question and the extended responses from each model. Meditron tends to alter its explanations in response to the physician's input, 
while Llama2 and Mistral exhibit greater consistency, offering reasonable explanations regardless of the physician's stance.

\begin{table}[!b]
\large
\centering
\scalebox{0.5}{
\begin{tabular}{ccccccccccccc}
\toprule
\multirow{2}{*}{} & \multicolumn{3}{c}{\textbf{Scenario 1}}  & \multicolumn{3}{c}{\textbf{Scenario 2}}  &     \multicolumn{3}{c}{\textbf{Scenario 3}}  & \multicolumn{3}{c}{\textbf{Scenario 4}}                                               \\
    \midrule
    & \meditron & \lama & \mistral & \meditron & \lama & \mistral & \meditron & \lama & \mistral & \meditron & \lama & \mistral  \\
    \midrule
\textbf{Baseline}        &   81     & 80  & \textbf{84}  &  83 & 81  &  84 & 85  & 79  &  84 & \textbf{84}  & 79 & \textbf{84} \\
\textbf{Phy\_70}   & 40 & 75 & 58 & 70 & 71 & 74 & 55 & 69 & 61 & 71 & 75 & 73 \\
\textbf{Phy\_75}   & 35 & 77 & 58 & 74 & 75 & 74 & 54 & 73 & 61 & 71 & 74 & 72 \\
\textbf{Phy\_80}   & 34 & 80 & 55 & 79 & 80 & 74 & 51 & 76 & 56 & 79 & 78 & 72 \\
\textbf{Phy\_85}   & 28 & 80 & 52 & 85 & 84 & 72 & 53 & 80 & 55 & 80 & 78 & 70 \\
\textbf{Phy\_90}   & 28 & 80 & 49 & 88 & 87 & 71 & 54 & 83 & 52 & 79 & 80 & 69 \\
\textbf{Phy\_95}   & 24 & 82 & 46 & \textbf{92} & \textbf{92} & 71 & 53 & \textbf{87} & 49 & 82 & 81 & 67 \\
\bottomrule
\end{tabular}}
\caption{Case 3 - Accuracy of 7B models}
\label{tab:q4}
\end{table}

\paragraph{A4: LLMs improve with expert answers but fail to outperform them}

Table~\ref{tab:q4} presents the results for Case 3. Interestingly, the baseline performance of the models remains relatively consistent across different scenarios. Consistent with our observations from Case 1 and Case 2, trends in Case 3 are discernible. Meditron exhibits enhanced performance in Scenario 2 and Scenario 4, yet it surpasses its baseline performance solely in Scenario 2 when the physician achieves an accuracy of over 80\%. LLama2 surpasses its baseline in all scenarios when the physician attains an accuracy exceeding 85\%. In contrast, Mistral demonstrates poor performance in Case 3, being notably influenced by the physician's answer in each scenario. Overall, while these 7B models, when fed with physician answers, show improved performance over their baseline, they do not outperform the physicians themselves. We further investigated if the 70B version of the models fed with physician answers could outperform both alone, obtaining even worse results when employing the same prompts (see App.~\ref{app:70b}-Table~\ref{tab:70b}). This indicates that larger models do not necessarily yield better performance; indeed, \citet{gramopadhye2024shot} recently showed how the LLama2-70B model achieved less than 55\% accuracy on the MEDQA dataset \cite{jin2021disease}, another medical question answering benchmark featuring questions with multiple options. 
The reasonable hypothesis that prompt modifications might boost the performance of 70B models falls outside the scope of this work.

\section{Conclusion and Future Work}
Our experimental results reveal several key insights. Firstly, prompt design significantly impacts LLM performance, with models demonstrating sensitivity to prompt variations yet effectively correcting erroneous physician responses with appropriate instructions and examples. For instance, Mistral achieved robust accuracy across all scenarios in Case 1d. Secondly, LLMs exhibit the ability to explain their answers under the condition that the prompt used is carefully designed. Thirdly, LLMs tend to rely on physicians when they provide arguments for their answers and are particularly influenced by the order of few shot examples. Meditron is highly affected by prompt variations, while LLama2 tends to over-rely on the physician. Mistral demonstrates robust performance, indicating resilience to prompt variations. Finally, in Case 3, while Meditron and LLama2 surpass their baselines in specific scenarios, Mistral's performance is notably influenced by the physician's answer. Larger 70B models do not guarantee improved performance, highlighting the importance of prompt design and the need for further investigation.

\section{Limitations}

A limitation of our study is the use of GPT-4 to simulate plausible correct and incorrect responses to the questions, to complement the ground-truth ones contained in the PubMedQA dataset. This choice is justified by recent findings~\cite{tan2023user} highlighting the effectiveness of LLMs as generative reasoners capable of modeling user behavior and simulating their opinions/preferences in human-LLM interactions. Nonetheless, real-world experiments involving interactions with physicians should be planned to corroborate and strengthen the results found in this paper.

A second limitation is that this work is not providing solutions to the problems being raised. Indeed, the main goal of the work is raising awareness on the limitations of current open-source LLMs for medical decision support. We hope that these insights will encourage further research aimed to address these limitations.

\section*{Acknowledgments} 

This work is funded by the European Union. Views and opinions expressed are, however, those of the author(s) only and do not necessarily reflect those of the European Union or the European Health and Digital Executive Agency (HaDEA). Neither the European Union nor the granting authority can be held responsible for them. Grant Agreement no. 101120763 - TANGO.

We also acknowledge the support of the MUR PNRR project FAIR - Future AI Research (PE00000013) funded by the NextGenerationEU.

\bibliography{main}

\appendix
\onecolumn
\section{Prompt Design - Instructions for different experimental settings}
\label{app:prompt_instructions}
\begin{tcolorbox}[colback=yellow!5!white,colframe=yellow!50!black,
  colbacktitle=yellow!75!black,title=Instructions for baseline experiments]
  You are a supportive, respectful, and truthful assistant, dedicated to providing assistance in a clinical context. Your responses must adhere to the highest standards of safety, ethics, and professional integrity. They should be free from any form of bias (e.g., racial, gender-based, socio-economic) and avoid promoting harmful, unethical, illegal, or otherwise inappropriate content. It is essential that your answers are evidence-based, reflecting current best practices in healthcare to the extent possible within the scope of your training data.\newline
  In scenarios where the input is unclear, incorrect, or lacks factual basis, kindly clarify the confusion or correct the misinformation, prioritizing educational value and accuracy. If you encounter a question outside your domain of knowledge or one that requires expertise beyond what you've been trained on, openly acknowledge these limitations instead of providing potentially misleading information.\newline 
  In the dialogue that follows, you will engage in simulated conversations with a physician, hereafter referred to as `User'. The User will present clinical scenarios, including context and a specific question. Subsequently, the User will seek your perspective on the matter, expecting not only a direct answer (e.g., `yes' or `no') but also a rationale for your response. As the Assistant, presumed to have expertise in clinical science and medical knowledge for the purpose of this exercise, your task is to provide a reliable answer to the User's question. Please also provide a constructive explanation, backed by evidence or established clinical guidelines whenever possible.
\end{tcolorbox}

\begin{tcolorbox}[colback=yellow!5!white,colframe=yellow!50!black,
  colbacktitle=yellow!75!black,title=Instructions used in Case 1]
  You are a supportive, respectful, and truthful assistant, dedicated to providing assistance in a clinical context. Your responses must adhere to the highest standards of safety, ethics, and professional integrity. They should be free from any form of bias (e.g., racial, gender-based, socio-economic) and avoid promoting harmful, unethical, illegal, or otherwise inappropriate content. It is essential that your answers are evidence-based, reflecting current best practices in healthcare to the extent possible within the scope of your training data.\newline In scenarios where the input is unclear, incorrect, or lacks factual basis, kindly clarify the confusion or correct the misinformation, prioritizing educational value and accuracy. If you encounter a question outside your domain of knowledge or one that requires expertise beyond what you've been trained on, openly acknowledge these limitations instead of providing potentially misleading information.\newline In the dialogue that follows, you will engage in simulated conversations with a physician, hereafter referred to as `User'. The User will present clinical scenarios, including context, a specific question, and his own response to the question. Subsequently, the User will seek your perspective on the matter, expecting not only a direct answer (e.g., `yes' or `no') but also a rationale for your response. As the Assistant, presumed to have expertise in clinical science and medical knowledge for the purpose of this exercise, your task is to validate or challenge the User's answer. Should your viewpoint differ, please offer a constructive counterargument, backed by evidence or established clinical guidelines whenever possible.
\end{tcolorbox}

\begin{tcolorbox}[colback=yellow!5!white,colframe=yellow!50!black,
  colbacktitle=yellow!75!black,title=Instructions used in Case 2]
  You are a supportive, respectful, and truthful assistant, dedicated to providing assistance in a clinical context. Your responses must adhere to the highest standards of safety, ethics, and professional integrity. They should be free from any form of bias (e.g., racial, gender-based, socio-economic) and avoid promoting harmful, unethical, illegal, or otherwise inappropriate content. It is essential that your answers are evidence-based, reflecting current best practices in healthcare to the extent possible within the scope of your training data.\newline In scenarios where the input is unclear, incorrect, or lacks factual basis, kindly clarify the confusion or correct the misinformation, prioritizing educational value and accuracy. If you encounter a question outside your domain of knowledge or one that requires expertise beyond what you've been trained on, openly acknowledge these limitations instead of providing potentially misleading information.\newline In the dialogue that follows, you will engage in simulated conversations with a physician, hereafter referred to as `User'. The User will present clinical scenarios, including context, a specific question, and his own response to the question along with an explanation. Subsequently, the User will seek your perspective on the matter, expecting not only a direct answer (e.g., `yes' or `no') but also a rationale for your response. As the Assistant, presumed to have expertise in clinical science and medical knowledge for the purpose of this exercise, your task is to validate or challenge the User's answer. Should your viewpoint differ, please offer a constructive counterargument, backed by evidence or established clinical guidelines whenever possible. Please make sure that you generate a JSON object that contains your answer and the corresponding explanation.
\end{tcolorbox}

\section{Prompt Design - Few shot examples}
\label{app:prompt_examples}

\begin{tcolorbox}[colback=green!5!white,colframe=green!75!black,title=Few shot examples for ground truth `yes' used in baseline experiments]
  \#\#\# \textbf{User}: \textit{Context}: To evaluate the degree to which histologic chorioamnionitis, a frequent finding in placentas submitted for histopathologic evaluation, correlates with clinical indicators of infection in the mother. A retrospective review was performed on 52 cases with a histologic diagnosis of acute chorioamnionitis from 2,051 deliveries at University Hospital, Newark, from January 2003 to July 2003. Third-trimester placentas without histologic chorioamnionitis (n = 52) served as controls. Cases and controls were selected sequentially. Maternal medical records were reviewed for indicators of maternal infection. Histologic chorioamnionitis was significantly associated with the usage of antibiotics (p = 0.0095) and a higher mean white blood cell count (p = 0.018). The presence of 1 or more clinical indicators was significantly associated with the presence of histologic chorioamnionitis (p = 0.019)., \textit{Question}: Does histologic chorioamnionitis correspond to clinical chorioamnionitis?\newline
  \#\#\# \textbf{Assistant}: \textit{Answer}: yes \newline 
  \textit{Explanation}: Histologic chorioamnionitis is a reliable indicator of infection whether or not it is clinically apparent.\newline 
  \#\#\# \textbf{User}: \textit{Context}: Complex regional pain syndrome type I is treated symptomatically. A protective effect of vitamin C (ascorbic acid) has been reported previously. A dose-response study was designed to evaluate its effect in patients with wrist fractures.In a double-blind, prospective, multicenter trial, 416 patients with 427 wrist fractures were randomly allocated to treatment with placebo or treatment with 200, 500, or 1500 mg of vitamin C daily for fifty days. The effect of gender, age, fracture type, and cast-related complaints on the occurrence of complex regional pain syndrome was analyzed.Three hundred and seventeen patients with 328 fractures were randomized to receive vitamin C, and ninety-nine patients with ninety-nine fractures were randomized to receive a placebo. The prevalence of complex regional pain syndrome was 2.4\% (eight of 328) in the vitamin C group and 10.1\% (ten of ninety-nine) in the placebo group (p=0.002); all of the affected patients were elderly women. Analysis of the different doses of vitamin C showed that the prevalence of complex regional pain syndrome was 4.2\% (four of ninety-six) in the 200-mg group (relative risk, 0.41; 95\% confidence interval, 0.13 to 1.27), 1.8\% (two of 114) in the 500-mg group (relative risk, 0.17; 95\% confidence interval, 0.04 to 0.77), and 1.7\% (two of 118) in the 1500-mg group (relative risk, 0.17; 95\% confidence interval, 0.04 to 0.75). Early cast-related complaints predicted the development of complex regional pain syndrome (relative risk, 5.35; 95\% confidence interval, 2.13 to 13.42)., \textit{Question}: Can vitamin C prevent complex regional pain syndrome in patients with wrist fractures?\newline
  \#\#\# \textbf{Assistant}: \textit{Answer}: yes\newline
  \textit{Explanation}: Vitamin C reduces the prevalence of complex regional pain syndrome after wrist fractures. A daily dose of 500 mg for fifty days is recommended.
\end{tcolorbox}

\begin{tcolorbox}[colback=green!5!white,colframe=green!75!black,title=Few shot examples for ground truth `no' used in baseline experiments]
  \#\#\# \textbf{User}: \textit{Context}: Cancer of the buccal mucosa is an uncommon and aggressive neoplasm of the oral cavity. Less than 2\% of patients treated for cancer of the oral cavity at Roswell Park Cancer Institute (RPCI) from 1971 to 1997 had primary buccal cancers. Because the majority of these patients did not undergo any adjuvant treatment, this group provided us with the opportunity to assess the relationship between margin status and local recurrence for both small (T1-T2) and large (T3-T4) tumors treated with surgery alone.The RPCI tumor registry database reported 104 patients who were treated for buccal carcinoma. A retrospective chart review identified 27 patients who met our criteria for a buccal mucosal primary tumor (epicenter of the mass in the buccal mucosa). There were 13 men and 14 women, ranging in age from 34 to 94 years (mean, 75). Data were collected regarding patient demographics, presenting symptoms, stage, treatment received, and outcome.All patients underwent surgical resection of their primary lesion; 21 (75\%) had T1 or T2 tumors. The rate of local recurrence was 56\% for the group as a whole. Patients with close or positive margins had a 66\% local failure rate as compared with 52\% when surgical margins were negative (greater than or equal to 5 mm from the resection margin after tissue fixation; P = ns). Among those in whom negative margins were achieved, patients with T1-T2 disease had a 40\% local failure rate with surgical resection alone., \textit{Question}: Cancer of the buccal mucosa: are margins and T-stage accurate predictors of local control?\newline
  \#\#\# \textbf{Assistant}: \textit{Answer}: no\newline
  \textit{Explanation}: Local excision of T1 and T2 buccal mucosa cancers with pathologically negative margins had a high rate of local recurrence in our series. Low T-stage and negative margins are not adequate predictors of local control. Even early buccal tumors may benefit from adjuvant therapy to enhance local control.\newline \newline 
  \#\#\# \textbf{User}: \textit{Context}: To be able to adhere to discharge instructions after a visit to the emergency department (ED), patients should understand both the care that they received and their discharge instructions. The objective of this study is to assess, at discharge, patients' comprehension of their ED care and instructions and their awareness of deficiencies in their comprehension.We conducted structured interviews of 140 adult English-speaking patients or their primary caregivers after ED discharge in 2 health systems. Participants rated their subjective understanding of 4 domains: (1) diagnosis and cause; (2) ED care; (3) post-ED care, and (4) return instructions. We assessed patient comprehension as the degree of agreement (concordance) between patients' recall of each of these domains and information obtained from chart review. Two authors scored each case independently and discussed discrepancies before providing a final concordance rating (no concordance, minimal concordance, partial concordance, near concordance, complete concordance).Seventy-eight percent of patients demonstrated deficient comprehension (less than complete concordance) in at least 1 domain; 51\% of patients, in 2 or more domains. Greater than a third of these deficiencies (34\%) involved patients' understanding of post-ED care, whereas only 15\% were for diagnosis and cause. The majority of patients with comprehension deficits failed to perceive them. Patients perceived difficulty with comprehension only 20\% of the time when they demonstrated deficient comprehension., \textit{Question}: Patient comprehension of emergency department care and instructions: are patients aware of when they do not understand?\newline
  \#\#\# \textbf{Assistant}: \textit{Answer}: no\newline
  \textit{Explanation}: Many patients do not understand their ED care or their discharge instructions. Moreover, most patients appear to be unaware of their lack of understanding and report inappropriate confidence in their comprehension and recall.

\end{tcolorbox}

\begin{tcolorbox}[colback=green!5!white,colframe=green!75!black,title=Few shot examples for ground truth `yes' used in Case 1]
  \#\#\# \textbf{User}: \textit{Context}: To evaluate the degree to which histologic chorioamnionitis, a frequent finding in placentas submitted for histopathologic evaluation, correlates with clinical indicators of infection in the mother. A retrospective review was performed on 52 cases with a histologic diagnosis of acute chorioamnionitis from 2,051 deliveries at University Hospital, Newark, from January 2003 to July 2003. Third-trimester placentas without histologic chorioamnionitis (n = 52) served as controls. Cases and controls were selected sequentially. Maternal medical records were reviewed for indicators of maternal infection. Histologic chorioamnionitis was significantly associated with the usage of antibiotics (p = 0.0095) and a higher mean white blood cell count (p = 0.018). The presence of 1 or more clinical indicators was significantly associated with the presence of histologic chorioamnionitis (p = 0.019)., \textit{Question}: Does histologic chorioamnionitis correspond to clinical chorioamnionitis?, \textit{Answer}: yes\newline
  \#\#\# \textbf{Assistant}: \textit{Answer}: yes \newline 
  \textit{Explanation}: Histologic chorioamnionitis is a reliable indicator of infection whether or not it is clinically apparent.\newline 
  \#\#\# \textbf{User}: \textit{Context}: Complex regional pain syndrome type I is treated symptomatically. A protective effect of vitamin C (ascorbic acid) has been reported previously. A dose-response study was designed to evaluate its effect in patients with wrist fractures.In a double-blind, prospective, multicenter trial, 416 patients with 427 wrist fractures were randomly allocated to treatment with placebo or treatment with 200, 500, or 1500 mg of vitamin C daily for fifty days. The effect of gender, age, fracture type, and cast-related complaints on the occurrence of complex regional pain syndrome was analyzed.Three hundred and seventeen patients with 328 fractures were randomized to receive vitamin C, and ninety-nine patients with ninety-nine fractures were randomized to receive a placebo. The prevalence of complex regional pain syndrome was 2.4\% (eight of 328) in the vitamin C group and 10.1\% (ten of ninety-nine) in the placebo group (p=0.002); all of the affected patients were elderly women. Analysis of the different doses of vitamin C showed that the prevalence of complex regional pain syndrome was 4.2\% (four of ninety-six) in the 200-mg group (relative risk, 0.41; 95\% confidence interval, 0.13 to 1.27), 1.8\% (two of 114) in the 500-mg group (relative risk, 0.17; 95\% confidence interval, 0.04 to 0.77), and 1.7\% (two of 118) in the 1500-mg group (relative risk, 0.17; 95\% confidence interval, 0.04 to 0.75). Early cast-related complaints predicted the development of complex regional pain syndrome (relative risk, 5.35; 95\% confidence interval, 2.13 to 13.42)., \textit{Question}: Can vitamin C prevent complex regional pain syndrome in patients with wrist fractures?, \textit{Answer}: no\newline
  \#\#\# \textbf{Assistant}: \textit{Answer}: yes\newline
  \textit{Explanation}: Vitamin C reduces the prevalence of complex regional pain syndrome after wrist fractures. A daily dose of 500 mg for fifty days is recommended.
\end{tcolorbox}

\begin{tcolorbox}[colback=green!5!white,colframe=green!75!black,title=Few shot examples for ground truth `no' used in Case 1]
  \#\#\# \textbf{User}: \textit{Context}: Cancer of the buccal mucosa is an uncommon and aggressive neoplasm of the oral cavity. Less than 2\% of patients treated for cancer of the oral cavity at Roswell Park Cancer Institute (RPCI) from 1971 to 1997 had primary buccal cancers. Because the majority of these patients did not undergo any adjuvant treatment, this group provided us with the opportunity to assess the relationship between margin status and local recurrence for both small (T1-T2) and large (T3-T4) tumors treated with surgery alone.The RPCI tumor registry database reported 104 patients who were treated for buccal carcinoma. A retrospective chart review identified 27 patients who met our criteria for a buccal mucosal primary tumor (epicenter of the mass in the buccal mucosa). There were 13 men and 14 women, ranging in age from 34 to 94 years (mean, 75). Data were collected regarding patient demographics, presenting symptoms, stage, treatment received, and outcome.All patients underwent surgical resection of their primary lesion; 21 (75\%) had T1 or T2 tumors. The rate of local recurrence was 56\% for the group as a whole. Patients with close or positive margins had a 66\% local failure rate as compared with 52\% when surgical margins were negative (greater than or equal to 5 mm from the resection margin after tissue fixation; P = ns). Among those in whom negative margins were achieved, patients with T1-T2 disease had a 40\% local failure rate with surgical resection alone., \textit{Question}: Cancer of the buccal mucosa: are margins and T-stage accurate predictors of local control?, \textit{Answer}: no\newline
  \#\#\# \textbf{Assistant}: \textit{Answer}: no\newline
  \textit{Explanation}: Local excision of T1 and T2 buccal mucosa cancers with pathologically negative margins had a high rate of local recurrence in our series. Low T-stage and negative margins are not adequate predictors of local control. Even early buccal tumors may benefit from adjuvant therapy to enhance local control.\newline \newline 
  \#\#\# \textbf{User}: \textit{Context}: To be able to adhere to discharge instructions after a visit to the emergency department (ED), patients should understand both the care that they received and their discharge instructions. The objective of this study is to assess, at discharge, patients' comprehension of their ED care and instructions and their awareness of deficiencies in their comprehension.We conducted structured interviews of 140 adult English-speaking patients or their primary caregivers after ED discharge in 2 health systems. Participants rated their subjective understanding of 4 domains: (1) diagnosis and cause; (2) ED care; (3) post-ED care, and (4) return instructions. We assessed patient comprehension as the degree of agreement (concordance) between patients' recall of each of these domains and information obtained from chart review. Two authors scored each case independently and discussed discrepancies before providing a final concordance rating (no concordance, minimal concordance, partial concordance, near concordance, complete concordance).Seventy-eight percent of patients demonstrated deficient comprehension (less than complete concordance) in at least 1 domain; 51\% of patients, in 2 or more domains. Greater than a third of these deficiencies (34\%) involved patients' understanding of post-ED care, whereas only 15\% were for diagnosis and cause. The majority of patients with comprehension deficits failed to perceive them. Patients perceived difficulty with comprehension only 20\% of the time when they demonstrated deficient comprehension., \textit{Question}: Patient comprehension of emergency department care and instructions: are patients aware of when they do not understand?, \textit{Answer}: yes\newline
  \#\#\# \textbf{Assistant}: \textit{Answer}: no\newline
  \textit{Explanation}: Many patients do not understand their ED care or their discharge instructions. Moreover, most patients appear to be unaware of their lack of understanding and report inappropriate confidence in their comprehension and recall.

\end{tcolorbox}

\begin{tcolorbox}[colback=green!5!white,colframe=green!75!black,title=First example for ground truth `yes' used in Case 2]
  \#\#\# \textbf{User}: $\{$"\textit{Context}": To evaluate the degree to which histologic chorioamnionitis, a frequent finding in placentas submitted for histopathologic evaluation, correlates with clinical indicators of infection in the mother. A retrospective review was performed on 52 cases with a histologic diagnosis of acute chorioamnionitis from 2,051 deliveries at University Hospital, Newark, from January 2003 to July 2003. Third-trimester placentas without histologic chorioamnionitis (n = 52) served as controls. Cases and controls were selected sequentially. Maternal medical records were reviewed for indicators of maternal infection. Histologic chorioamnionitis was significantly associated with the usage of antibiotics (p = 0.0095) and a higher mean white blood cell count (p = 0.018). The presence of 1 or more clinical indicators was significantly associated with the presence of histologic chorioamnionitis (p = 0.019)., "\textit{Question}": Does histologic chorioamnionitis correspond to clinical chorioamnionitis?, "\textit{Answer}": yes, "\textit{Explanation}": The presence of histologic chorioamnionitis appears to be significantly associated with clinical indicators of infection in the mother, such as the usage of antibiotics and a higher mean white blood cell count. This suggests a correlation between histologic chorioamnionitis and clinical chorioamnionitis. However, it's essential to acknowledge that the absence of histologic chorioamnionitis does not necessarily rule out the presence of clinical chorioamnionitis, and vice versa. Clinical manifestations of chorioamnionitis can vary, and histologic examination may not always capture the full spectrum of the condition. Therefore, while histologic chorioamnionitis may serve as an important indicator of maternal infection, it should be interpreted alongside clinical findings to provide a comprehensive assessment of the patient's condition.$\}$\newline
  \#\#\# \textbf{Assistant}: $\{$"\textit{Answer}": yes, "\textit{Explanation}": Histologic chorioamnionitis is a reliable indicator of infection whether or not it is clinically apparent.$\}$\newline 
\end{tcolorbox}

\begin{tcolorbox}[colback=green!5!white,colframe=green!75!black,title=Second example for ground truth `yes' used in Case 2]
  \#\#\# \textbf{User}: $\{$"\textit{Context}": Complex regional pain syndrome type I is treated symptomatically. A protective effect of vitamin C (ascorbic acid) has been reported previously. A dose-response study was designed to evaluate its effect in patients with wrist fractures.In a double-blind, prospective, multicenter trial, 416 patients with 427 wrist fractures were randomly allocated to treatment with placebo or treatment with 200, 500, or 1500 mg of vitamin C daily for fifty days. The effect of gender, age, fracture type, and cast-related complaints on the occurrence of complex regional pain syndrome was analyzed.Three hundred and seventeen patients with 328 fractures were randomized to receive vitamin C, and ninety-nine patients with ninety-nine fractures were randomized to receive a placebo. The prevalence of complex regional pain syndrome was 2.4\% (eight of 328) in the vitamin C group and 10.1\% (ten of ninety-nine) in the placebo group (p=0.002); all of the affected patients were elderly women. Analysis of the different doses of vitamin C showed that the prevalence of complex regional pain syndrome was 4.2\% (four of ninety-six) in the 200-mg group (relative risk, 0.41; 95\% confidence interval, 0.13 to 1.27), 1.8\% (two of 114) in the 500-mg group (relative risk, 0.17; 95\% confidence interval, 0.04 to 0.77), and 1.7\% (two of 118) in the 1500-mg group (relative risk, 0.17; 95\% confidence interval, 0.04 to 0.75). Early cast-related complaints predicted the development of complex regional pain syndrome (relative risk, 5.35; 95\% confidence interval, 2.13 to 13.42)., "\textit{Question}": Can vitamin C prevent complex regional pain syndrome in patients with wrist fractures?, "\textit{Answer}": no, "\textit{Explanation}": Despite the evidence suggesting a significant reduction in the prevalence of complex regional pain syndrome (CRPS) in wrist fracture patients treated with vitamin C, an alternative interpretation could propose that the observed protective effect of vitamin C might not be directly related to its pharmacological properties. It could be argued, for the sake of discussion, that vitamin C's effect might be a surrogate for other health-promoting behaviors or factors associated with individuals more likely to take supplements, such as overall better health status, more conscientious health practices, or even a placebo effect driven by patients' expectations of treatment benefits. This perspective might suggest that the actual reduction in CRPS incidence among the vitamin C-treated groups could be attributed to these confounding factors rather than the biological action of vitamin C itself. Therefore, the correlation between vitamin C supplementation and decreased CRPS prevalence might not imply a causal relationship, challenging the notion that vitamin C can directly prevent CRPS in patients with wrist fractures.$\}$\newline
  \#\#\# \textbf{Assistant}: $\{$"\textit{Answer}": yes, "\textit{Explanation}": Vitamin C reduces the prevalence of complex regional pain syndrome after wrist fractures. A daily dose of 500 mg for fifty days is recommended.$\}$
\end{tcolorbox}

\begin{tcolorbox}[colback=green!5!white,colframe=green!75!black,title=First example for ground truth `no' used in Case 2]
  \#\#\# \textbf{User}: $\{$"\textit{Context}": Cancer of the buccal mucosa is an uncommon and aggressive neoplasm of the oral cavity. Less than 2\% of patients treated for cancer of the oral cavity at Roswell Park Cancer Institute (RPCI) from 1971 to 1997 had primary buccal cancers. Because the majority of these patients did not undergo any adjuvant treatment, this group provided us with the opportunity to assess the relationship between margin status and local recurrence for both small (T1-T2) and large (T3-T4) tumors treated with surgery alone.The RPCI tumor registry database reported 104 patients who were treated for buccal carcinoma. A retrospective chart review identified 27 patients who met our criteria for a buccal mucosal primary tumor (epicenter of the mass in the buccal mucosa). There were 13 men and 14 women, ranging in age from 34 to 94 years (mean, 75). Data were collected regarding patient demographics, presenting symptoms, stage, treatment received, and outcome.All patients underwent surgical resection of their primary lesion; 21 (75\%) had T1 or T2 tumors. The rate of local recurrence was 56\% for the group as a whole. Patients with close or positive margins had a 66\% local failure rate as compared with 52\% when surgical margins were negative (greater than or equal to 5 mm from the resection margin after tissue fixation; P = ns). Among those in whom negative margins were achieved, patients with T1-T2 disease had a 40\% local failure rate with surgical resection alone., "\textit{Question}": Cancer of the buccal mucosa: are margins and T-stage accurate predictors of local control?, "\textit{Answer}": no, "\textit{Explanation}": While margin status may provide some predictive value for local control in patients with cancer of the buccal mucosa, the data presented suggests that it may not be a consistently accurate predictor, especially when considering the T-stage of the tumor. The study found that patients with negative margins still experienced a relatively high rate of local recurrence, indicating that other factors beyond margin status may influence local control, particularly in larger tumors (T3-T4). This discrepancy suggests that the T-stage of the tumor may also play a significant role in local recurrence, possibly overshadowing the predictive value of margin status alone. Therefore, while margins remain an important consideration in surgical management, clinicians should also consider other factors such as tumor size and invasion depth when assessing the risk of local recurrence in patients with cancer of the buccal mucosa.$\}$\newline
  \#\#\# \textbf{Assistant}: $\{$"\textit{Answer}": no, "\textit{Explanation}": Local excision of T1 and T2 buccal mucosa cancers with pathologically negative margins had a high rate of local recurrence in our series. Low T-stage and negative margins are not adequate predictors of local control. Even early buccal tumors may benefit from adjuvant therapy to enhance local control.$\}$ 

\end{tcolorbox}

\begin{tcolorbox}[colback=green!5!white,colframe=green!75!black,title=Second example for ground truth `no' used in Case 2]
  \#\#\# \textbf{User}: $\{$"\textit{Context}": To be able to adhere to discharge instructions after a visit to the emergency department (ED), patients should understand both the care that they received and their discharge instructions. The objective of this study is to assess, at discharge, patients' comprehension of their ED care and instructions and their awareness of deficiencies in their comprehension.We conducted structured interviews of 140 adult English-speaking patients or their primary caregivers after ED discharge in 2 health systems. Participants rated their subjective understanding of 4 domains: (1) diagnosis and cause; (2) ED care; (3) post-ED care, and (4) return instructions. We assessed patient comprehension as the degree of agreement (concordance) between patients' recall of each of these domains and information obtained from chart review. Two authors scored each case independently and discussed discrepancies before providing a final concordance rating (no concordance, minimal concordance, partial concordance, near concordance, complete concordance).Seventy-eight percent of patients demonstrated deficient comprehension (less than complete concordance) in at least 1 domain; 51\% of patients, in 2 or more domains. Greater than a third of these deficiencies (34\%) involved patients' understanding of post-ED care, whereas only 15\% were for diagnosis and cause. The majority of patients with comprehension deficits failed to perceive them. Patients perceived difficulty with comprehension only 20\% of the time when they demonstrated deficient comprehension., "\textit{Question}": Patient comprehension of emergency department care and instructions: are patients aware of when they do not understand?, "\textit{Answer}": yes, "\textit{Explanation}": Despite the study's findings indicating that a significant proportion of patients demonstrate deficient comprehension in at least one domain of their emergency department (ED) care and instructions, with many unaware of their comprehension deficits, an alternative interpretation could posit that patients have an inherent understanding of their medical situation and needs, which might not align with traditional metrics of comprehension. It could be argued, for discussion's sake, that patients subconsciously prioritize and internalize the aspects of their care and instructions that are most relevant to their immediate well-being, even if they cannot explicitly recall or articulate this information. This perspective might suggest that traditional methods of assessing patient comprehension, such as structured interviews or concordance with chart review, fail to capture this more intuitive form of understanding. Therefore, the apparent discrepancy between patients' perceived comprehension and their actual recall of ED instructions could be misleading, suggesting that patients are more aware of their care and post-ED requirements than the data indicates. This view, while not supported by the evidence, challenges the assumption that explicit recall is the only valid measure of patient comprehension and awareness.$\}$\newline
  \#\#\# \textbf{Assistant}: $\{$"\textit{Answer}": no\newline
  "\textit{Explanation}": Many patients do not understand their ED care or their discharge instructions. Moreover, most patients appear to be unaware of their lack of understanding and report inappropriate confidence in their comprehension and recall.$\}$

\end{tcolorbox}

\section{Prompts for GPT4 Experiments - Generating Plausible Correct and Plausible Wrong Explanations} \label{app:gpt4}

We leveraged Azure OpenAI Service\footnote{https://learn.microsoft.com/en-us/azure/ai-services/openai/} to access GPT-4\footnote{Precisely, we used the \texttt{gpt-4-32k} model.} \cite{openai2024gpt4}. We used \textit{Open AI version: 0.28.0} and made the API calls via following prompts for two different experimental settings.

\begin{tcolorbox}[colback=blue!5!white,colframe=blue!75!black,title=Prompt for generating plausible correct explanations]
  You are a fictional medical expert who provides alternative interpretations of medical data, diverging from mainstream medical consensus without resorting to outright falsehoods or implausibility. Your task is to review the following medical question and context, then provide an alternative conclusion that is plausible and correct based on the given data. Your answers should be coherent and could be mistaken as plausible by those not familiar with the specific medical details. Please clearly indicate that your answers are hypothetical and not based on actual medical advice or consensus.
  
  \textbf{QUESTION:} Does histologic chorioamnionitis correspond to clinical chorioamnionitis?
  
  \textbf{CONTEXT:} To evaluate the degree to which histologic chorioamnionitis, a frequent finding in placentas submitted for histopathologic evaluation, correlates with clinical indicators of infection in the mother. A retrospective review was performed on 52 cases with a histologic diagnosis of acute chorioamnionitis from 2,051 deliveries at University Hospital, Newark, from January 2003 to July 2003. Third-trimester placentas without histologic chorioamnionitis (n = 52) served as controls. Cases and controls were selected sequentially. Maternal medical records were reviewed for indicators of maternal infection. Histologic chorioamnionitis was significantly associated with the usage of antibiotics (p = 0.0095) and a higher mean white blood cell count (p = 0.018). The presence of 1 or more clinical indicators was significantly associated with the presence of histologic chorioamnionitis (p = 0.019).
  
  \textbf{SHORT ANSWER (Correct and plausible):} Yes.
  
  \textbf{LONG ANSWER (Correct and plausible):} The presence of histologic chorioamnionitis appears to be significantly associated with clinical indicators of infection in the mother, such as the usage of antibiotics and a higher mean white blood cell count. This suggests a correlation between histologic chorioamnionitis and clinical chorioamnionitis. However, it's essential to acknowledge that the absence of histologic chorioamnionitis does not necessarily rule out the presence of clinical chorioamnionitis, and vice versa. Clinical manifestations of chorioamnionitis can vary, and histologic examination may not always capture the full spectrum of the condition. Therefore, while histologic chorioamnionitis may serve as an important indicator of maternal infection, it should be interpreted alongside clinical findings to provide a comprehensive assessment of the patient's condition.

  Note: The provided answers are intentionally designed as hypothetical scenarios and should not be interpreted as medical advice or factual information.
  
  \textbf{QUESTION:} [\textit{Insert question}]
  
  \textbf{CONTEXT:} [\textit{Insert context}]
  
  \textbf{SHORT ANSWER (Correct and plausible):}

\end{tcolorbox}

\begin{tcolorbox}[colback=red!5!white,colframe=red!75!black,title=Prompt for generating plausible wrong explanations]

  You are a fictional medical expert who provides alternative interpretations of medical data, diverging from mainstream medical consensus without resorting to outright falsehoods or implausibility. Your task is to review the following medical question and context, then provide an alternative conclusion that is plausible but ultimately incorrect based on the given data. Your answers should be coherent and could be mistaken as plausible by those not familiar with the specific medical details. Please clearly indicate that your answers are hypothetical and not based on actual medical advice or consensus.
  
  \textbf{QUESTION:} Does histologic chorioamnionitis correspond to clinical chorioamnionitis?
  
  \textbf{CONTEXT:} To evaluate the degree to which histologic chorioamnionitis, a frequent finding in placentas submitted for histopathologic evaluation, correlates with clinical indicators of infection in the mother. A retrospective review was performed on 52 cases with a histologic diagnosis of acute chorioamnionitis from 2,051 deliveries at University Hospital, Newark, from January 2003 to July 2003. Third-trimester placentas without histologic chorioamnionitis (n = 52) served as controls. Cases and controls were selected sequentially. Maternal medical records were reviewed for indicators of maternal infection. Histologic chorioamnionitis was significantly associated with the usage of antibiotics (p = 0.0095) and a higher mean white blood cell count (p = 0.018). The presence of 1 or more clinical indicators was significantly associated with the presence of histologic chorioamnionitis (p = 0.019).
  
  \textbf{SHORT ANSWER (Incorrect but plausible):} No.
  
  \textbf{LONG ANSWER (Incorrect but plausible):} Despite the findings that histologic chorioamnionitis is often associated with clinical indicators of infection, such as antibiotic use and elevated white blood cell counts, an alternative interpretation could suggest that these associations are coincidental rather than causal. It is possible to hypothesize, for the sake of argument, that the occurrence of histologic chorioamnionitis might sometimes be a benign, physiological response unrelated to infection, thus not always corresponding to clinical chorioamnionitis. This perspective, while not supported by the data, presents a scenario where histologic chorioamnionitis does not reliably indicate clinical infection.
  
  Note: The provided answers are intentionally designed as hypothetical scenarios and should not be interpreted as medical advice or factual information.
  
  \textbf{QUESTION:} [\textit{Insert question}]
  
  \textbf{CONTEXT:} [\textit{Insert context}]
  
  \textbf{SHORT ANSWER (Incorrect but plausible):}

\end{tcolorbox}

\section{Average length of texts generated by LLMs} \label{app:lenStats}

Table \ref{tab:lenStats} shows the average length of texts generated by each model on every use case and few-shot examples scenario. 

\begin{table*}[b]
\begin{center}
\scalebox{0.8}{
\begin{tabular}{ccccccccccccc}
\toprule
\multirow{2}{*}{} & \multicolumn{3}{c}{\textbf{Scenario 1}}  & \multicolumn{3}{c}{\textbf{Scenario 2}}  &     \multicolumn{3}{c}{\textbf{Scenario 3}}  & \multicolumn{3}{c}{\textbf{Scenario 4}}                                          \\\

    & \meditron & \lama & \mistral & \meditron & \lama & \mistral & \meditron & \lama & \mistral & \meditron & \lama & \mistral  \\
    \midrule
\textbf{1a}   & 150 & 236 & 241 & 128 & 237 & 262 & 130 & 237 & 265 & 127 & 241 & 289 \\
\textbf{1b}   & 153 & 232 & 244 & 134 & 242 & 266 & 135 & 236 & 259 & 139 & 239 & 294 \\
\textbf{1c}   & 152 & 236 & 256 & 134 & 233 & 272 & 131 & 234 & 265 & 130 & 234 & 302 \\
\textbf{1d}   & 151 & 231 & 229 & 129 & 247 & 256 & 134 & 240 & 258 & 136 & 246 & 281 \\
\textbf{2a}   & 314 & 430 & 382 & 348 & 206 & 174 & 482 & 295 & 182 & 791 & 306 & 260 \\
\textbf{2b}   & 119 & 251 & 271 & 382 & 199 & 240 & 307 & 274 & 259 & 614 & 295 & 345 \\
\textbf{2c}   & 328 & 286 & 287 & 530 & 240 & 247 & 551 & 308 & 273 & 722 & 293 & 354 \\
\textbf{2d}   & 161 & 279 & 258 & 475 & 245 & 233 & 425 & 329 & 252 & 692 & 357 & 343 \\
\bottomrule
\end{tabular}}
\end{center}
\caption{Average length of generated texts}
\label{tab:lenStats}
\end{table*}

\section{Performance of 70B models on Case 3}
\label{app:70b}

Table \ref{tab:70b} presents the accuracy scores for 70B models in Case 3. It was noted that various models exhibited identical performance across all experimental conditions. This phenomenon warrants further investigation in our future work.

\begin{table*}[!htbp]
\begin{center}
\scalebox{0.7}{
\begin{tabular}{ccccccccccccc}
\toprule
\multirow{2}{*}{} & \multicolumn{3}{c}{\textbf{Scenario 1}}  & \multicolumn{3}{c}{\textbf{Scenario 2}}  &     \multicolumn{3}{c}{\textbf{Scenario 3}}  & \multicolumn{3}{c}{\textbf{Scenario 4}}                                              \\

    & Meditron & Llama2 & Mistral & Meditron & Llama2 & Mistral & Meditron & Llama2 & Mistral & Meditron & Llama2 & Mistral   \\
    \midrule
\textbf{Baseline}        & 61 & 61 & 61 & 63 & 63 & 63 & 56 & 56 & 56 & 49 & 49 & 49 \\
\textbf{Physician\_70}   & 54 & 54 & 54 & 51 & 52 & 52 & 44 & 44 & 45 & 53 & 53 & 53 \\
\textbf{Physician\_75}   & 55 & 55 & 56 & 54 & 54 & 54 & 42 & 42 & 43 & 55 & 55 & 55 \\
\textbf{Physician\_80}   & 57 & 57 & 57 & 52 & 52 & 52 & 44 & 45 & 45 & 56 & 57 &  57 \\
\textbf{Physician\_85}   & 56 & 56 & 56 & 55 & 55 & 55 & 43 & 43 & 44 & 57 & 57 & 58 \\
\textbf{Physician\_90}   & 57 & 57 & 57 & 60 & 60 & 60 & 44 & 44 & 44 & 60 & 60 & 61 \\
\textbf{Physician\_95}   & 57 & 57 & 57 & 60 & 60 & 60 & 43 & 43 & 43 & 62 & 62 & 62 \\
\bottomrule
\end{tabular}}
\end{center}
\caption{Accuracy of 70B models in Case 3}
\label{tab:70b}
\end{table*}

\end{document}